\pdfoutput=1

\documentclass[runningheads]{llncs}

\usepackage[dvipsnames]{xcolor}
\colorlet{CONSTDIR}{YellowGreen}
\colorlet{LIMIT}{YellowOrange}
\colorlet{OBJNAME}{Aquamarine}
\colorlet{VAR}{Goldenrod}
\colorlet{PARAM}{Apricot}
\colorlet{OBJDIR}{GreenYellow}

\usepackage{tikz}
\usetikzlibrary{automata, positioning, shapes, shapes.multipart, arrows, arrows.meta, calc}
\tikzstyle{fancytitle}=[fill=black, text=white, font=\scshape]

\usepackage[frozencache, cachedir=minted-cache]{minted}
\usepackage{etoolbox}
\makeatletter
\patchcmd{\minted@colorbg}{\medskip}{}{}{}
\patchcmd{\endminted@colorbg}{\medskip}{}{}{}
\makeatother
\usepackage{amsmath}
\usepackage{hyperref}
\usepackage{multirow}
\usepackage{booktabs}
\usepackage[numbers, sort]{natbib}
\usepackage[]{inputenc}
\usepackage[T1]{fontenc}
\usepackage[absolute,overlay]{textpos}

\makeatletter 
\renewcommand\@biblabel[1]{#1.} 
\makeatother

\newcommand{\stos}{\textsc{Seq2Seq}}

\title{
Highlighting Named Entities in Input for Auto-Formulation of Optimization Problems
}
\titlerunning{Highlighting Named Entities for Auto-Formulation of Optimization Problems}

\author{Neeraj Gangwar \and Nickvash Kani}
\authorrunning{Gangwar and Kani}
\institute{
  Electrical and Computer Engineering\\
  University of Illinois Urbana-Champaign, IL, USA\\
  \email{\{gangwar2,kani\}@illinois.edu}
}

\begin{document}
\maketitle

\begin{textblock*}{\textwidth}(4.7cm, 25cm) 
   \footnotesize
   \noindent\textcolor{red}{\fbox{\parbox{\textwidth}{
   \noindent This preprint has not undergone peer review or any post-submission improvements or corrections. The Version of Record of this contribution is published in CICM 2023 (Lecture Notes in Computer Science, vol 14101, Springer Cham) and is available online at \href{https://doi.org/10.1007/978-3-031-42753-4_9}{https://doi.org/10.1007/978-3-031-42753-4\_9}. For the most up-to-date information, consult the Version of Record.
   }}}
\end{textblock*}

\begin{abstract}
    Operations research deals with modeling and solving real-world problems as mathematical optimization problems. While solving mathematical systems is accomplished by analytical software, formulating a problem as a set of mathematical operations has been typically done manually by domain experts. Recent machine learning methods have shown promise in converting textual problem descriptions to corresponding mathematical formulations. This paper presents an approach that converts linear programming word problems into mathematical formulations. We leverage the named entities in the input and augment the input to highlight these entities. Our approach achieves the highest accuracy among all submissions to the NL4Opt Competition, securing first place in the generation track.
\end{abstract}

\section{Introduction}
Operations research deals with modeling and solving real-world problems as mathematical optimization problems \citep{tao2020analytics, ma2016research, beairsto2021identifying, bitran2016overview}. There exist optimization solvers powered by efficient algorithms \citep{nash2000dantzig, karmarkar1984new} that can be used to solve these problems. However, these solvers do not directly take problem descriptions as input, and domain experts are required to model a problem into a mathematical formulation. \citet{ramamonjison2022augmenting} described an interactive system that can suggest a formulation based on the natural language description of a linear programming problem. Their system consists of two main components - an \textit{entity tagger} to tag named entities in the input problem description and a \textit{formulation generator} to generate the mathematical formulation. They also published a dataset of linear programming problems with two tasks - named entity tagging and mathematical formulation generation.

In this paper, \textit{we focus on the formulation generation}. The input to formulation generation consists of a word problem, labeled semantic entities, and the order mapping of variables. We propose a novel approach that leverages the labeled semantic entities. In particular, we highlight the named entities present in the problem description using XML-like start and end tags. Our results show that a \stos{} model, like BART \citep{lewis-etal-2020-bart}, can leverage this information while generating the mathematical formulation. Our approach achieves the highest accuracy among all submissions to the generation track of the NL4Opt Competition.\footnote{\href{https://nl4opt.github.io}{nl4opt.github.io}} Unlike the NL4Opt Generation dataset, labeled named entity information is not available in most applications. We show that these applications may still benefit by using an existing named entity system to label the input and highlight the identified named entities. Furthermore, we present an ablation study, highlighting the impact of different components of the model. Our source code is available on GitHub.\footnote{\href{https://github.com/mlpgroup/nl4opt-generation}{github.com/mlpgroup/nl4opt-generation}}

\begin{figure*}[!t]
    \centering
    \begin{tikzpicture}[font=\scriptsize\sffamily]
    \node [draw, text width=6cm, align=left] (INPUT) at (0, 0) {
        \begin{tabular}{p{0.97\linewidth}}
        \\
        {\bf\sffamily Problem:}\\
        {\setlength\fboxsep{2pt}
        A berry picker must pick \colorbox{CONSTDIR}{\strut {\tiny \bf \textsc{const\_dir}} at least} \colorbox{LIMIT}{\strut {\tiny \bf \textsc{limit}} 3000} strawberries and \colorbox{LIMIT}{\strut {\tiny \bf \textsc{limit}} 15000} raspberries. He visits two farms. For each \colorbox{OBJNAME}{\strut {\tiny \bf \textsc{obj\_name}} hour} at \colorbox{VAR}{\strut {\tiny \bf \textsc{var}} farm 1} he spends, he can pick \colorbox{PARAM}{\strut {\tiny \bf \textsc{param}} 50} strawberries and \colorbox{PARAM}{\strut {\tiny \bf \textsc{param}} 300} raspberries. For each \colorbox{OBJNAME}{\strut {\tiny \bf \textsc{obj\_name}} hour} at \colorbox{VAR}{\strut {\tiny \bf \textsc{var}} farm 2} he spends, he can catch \colorbox{PARAM}{\strut {\tiny \bf \textsc{param}} 70} strawberries and \colorbox{PARAM}{\strut {\tiny \bf \textsc{param}} 200} raspberries. How many \colorbox{OBJNAME}{\strut {\tiny \bf \textsc{obj\_name}} hours} should he spend at each farm to \colorbox{OBJDIR}{\strut {\tiny \bf \textsc{obj\_dir}} minimize} the \colorbox{OBJNAME}{\strut {\tiny \bf \textsc{obj\_name}} amount of time} he spends at both farms?
        } \\
        \\
        \textbf{Order Mapping:} \\
        farm 1: 0, farm 2: 1 \\
        \end{tabular}
    };
    
    \node [rectangle split, rectangle split parts=2, rectangle split draw splits=false, draw, text width=5cm, align=left, right=of INPUT, right=0.75cm] (OUTPUT) {
        \begin{minted}[tabsize=2, autogobble, bgcolor=white]{bash}
{
    'type': 'objvar',
    'direction': 'minimize',
    'name': 'amount of time',
    'vars': ['farm 2', 'farm 1']
}
            \end{minted}
            
            \nodepart{two}
            
            \begin{minted}[tabsize=2, autogobble, bgcolor=white]{bash}
{
    'type': 'linear',
    'direction': 'at least',
    'limit': '3000',
    'terms': {
        'farm 2': '70',
        'farm 1': '50'
    },
    'operator': 'GREATER_OR_EQUAL'
},
{
    'type': 'linear',
    'direction': 'at least',
    'limit': '15000',
    'terms': {
        'farm 1': '300',
        'farm 2': '200'
    },
    'operator': 'GREATER_OR_EQUAL'
}
            \end{minted}
        };

        \node [draw, text width=5cm, align=left, anchor=north west, below=of OUTPUT, minimum height=3cm] (ALGEBRAIC) {
            {\tiny \textbf{Objective:}}
            $$\min x + y$$
            
            {\tiny \textbf{Constraints:}}
            \begin{equation*}
                \begin{split}
                    50 x + 70 y &\ge 3000 \\
                    300 x + 200 y &\ge 15000
                \end{split}
            \end{equation*}
        };

        \node [draw, text width=6cm, align=left, left=of ALGEBRAIC, minimum height=3cm, left=0.75cm] (OBJTITLE) {
				
        };

        \node [rectangle split, rectangle split horizontal, rectangle split parts=2, draw, text width=0.75cm, align=center, below=-1cm] (OBJECTIVE) at (OBJTITLE) {
            $1$
            \nodepart{two}
            $1$
        };

        \node [rectangle split, rectangle split horizontal, rectangle split parts=2, text width=0.75cm, align=center, below=0.2cm] (OBJORDER) at (OBJECTIVE) {
            0
            \nodepart{two}
            1
        };
		
        \node [left=of OBJECTIVE, left=1.5cm, anchor=west] (OBJ) {\tiny \textbf{Objective:}};
		
        \node [rectangle split, rectangle split horizontal, rectangle split parts=2, draw, text width=0.75cm, align=center, below=0.8cm] (CONST1) at (OBJECTIVE) {
            $-50$
            \nodepart{two}
            $-70$
        };
			
        \node [left=of CONST1, left=1.5cm, anchor=west] {\tiny \textbf{Constraints:}};
		
        \node [rectangle split, rectangle split horizontal, rectangle split parts=2, draw, text width=0.75cm, align=center, below=0.4cm] (CONST2) at (CONST1) {
            $-300$
            \nodepart{two}
            $-200$
        };
		
        \node [rectangle split, rectangle split horizontal, rectangle split parts=2, text width=0.75cm, align=center, below=0.2cm] (CONSTORDER) at (CONST2) {
            0
            \nodepart{two}
            1
        };
		
        \node [draw, text width=1cm, align=center, right=1.25cm] (CONST1VAL) at (CONST1) {$-3000$};
		
        \node [draw, text width=1cm, align=center, right=1.25cm] (CONST2VAL) at (CONST2) {$-15000$};
    
        \draw [densely dashed] (OUTPUT.text split west) -- (OUTPUT.text split east);
    
        \node [rotate=90] at ($(OUTPUT.one east) - (3mm, 0)$) {\scriptsize \textsc{objective}};
        
        \node [rotate=90] at ($(OUTPUT.two east) - (3mm, 0)$) {\scriptsize \textsc{constraints}};
        
        \draw [double, -Stealth] (INPUT) edge (OUTPUT);
        
        \node [fancytitle, minimum width=1cm, anchor=west, right=2mm, font=\tiny\scshape\sffamily] at (INPUT.north west) {Input};
        
        \node [fancytitle, minimum width=1cm, anchor=west, font=\tiny\scshape\sffamily, right=2mm] at (OUTPUT.north west) {Output};

        \node [fancytitle, minimum width=1cm, anchor=west, right=2mm, font=\tiny\scshape\sffamily] at (OBJTITLE.north west) {Canonical};

        \node [fancytitle, minimum width=1cm, anchor=west, right=2mm, font=\tiny\scshape\sffamily] at (ALGEBRAIC.north west) {Algebraic};

        \draw [{Implies[length=6mm, width=6mm]}-{Implies[length=6mm, width=6mm]}, double, double distance=1mm] (OUTPUT.south) -- (ALGEBRAIC.north);

        \draw [{Implies[length=6mm, width=6mm]}-{Implies[length=6mm, width=6mm]}, double, double distance=1mm] (OBJTITLE.east) -- (ALGEBRAIC.west);
    \end{tikzpicture}
    \caption{\textit{(Top)} An example input-output pair from the dataset which consists of the problem statement, labeled named entities, and the order mapping of variable mentions. \textit{(Bottom)} The algebraic and the corresponding canonical formulations are shown on the right and left, respectively. We use the canonical formulation to represent the output in our implementation. To get the canonical formulation, it is assumed that the objective is always minimized, and the constraints are upper bounds. For a maximization objective or a lower bound constraint, the signs of the multipliers are inverted.}
    \label{fig:overview}
\end{figure*}
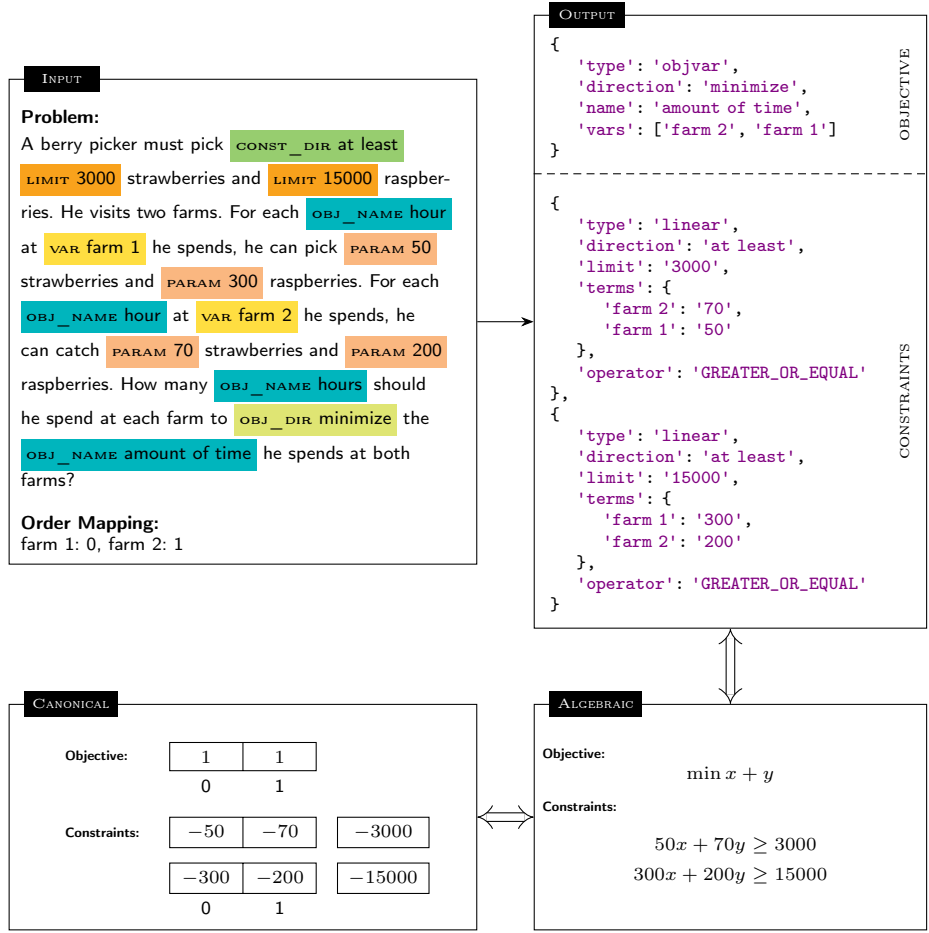

\section{Related Work}
The problem of identifying named entities in linear programming word problems and generating their mathematical formulation was proposed by \citet{ramamonjison2022augmenting}. In their work, they provided a baseline model for the formulation generation task, which used a two-step mapping approach. They used the BART-base model \citep{lewis-etal-2020-bart} with copy mechanism to generate an intermediate representation of the problem, which was then parsed into a canonical formulation. \citet{Jang2022TagEA} used entity tag embeddings with BART-large \citep{lewis-etal-2020-bart} to leverage the named entity information and introduced data augmentation in their approach. \citet{ning2023novel} used prompt-guided input along with adversarial learning in their approach. Their approach used BART-base and achieved competitive performance compared to approaches utilizing BART-large. On the other hand, \citet{He2022LinearPW} used multitask learning and input pre-processing. They augmented the input problem description by encapsulating the named entities in corresponding tags. Their approach used prompt-guided input and generated either the objective or a constraint at a time.

Our approach utilizes the named entity information and augments the input problem description by encapsulating the named entities in their corresponding tags. We use BART with copy mechanism to generate an intermediate representation, which is then parsed into a canonical formulation. Differing from \citet{He2022LinearPW}, \citet{ning2023novel}, and \citet{ramamonjison2022augmenting}, we use ``\textit{all-at-once}'' strategy and generate the objective and constraints at once. We do not use multitask or adversarial learning in our approach. Unlike \citet{Jang2022TagEA}, our approach does not have an additional hyperparameter. Despite its simplicity, our approach achieves better results than the more complex submissions, suggesting all that is needed for optimal performance is well-tagged data and a large model.

\section{Proposed Approach}
Named entities carry important semantic information. For generating mathematical formulations from linear programming word problems, the highlighted named entities can be utilized to form the objective and constraints for the input problem (Fig.~\ref{fig:overview}). We leverage this information and hypothesize that this additional information can help a \stos{} model in generating the mathematical formulation. In our proposed approach, we highlight the named entities in the problem description before passing it to a \stos{} model.

\subsubsection{Named Entity-Based Augmentation.}
\begin{figure}[t]
    \centering
    \begin{tikzpicture}[font=\sffamily\scriptsize]
        \node [draw, align=left, text width=0.95\columnwidth] (INPUT) {
            A berry picker must pick \colorbox{CONSTDIR}{\strut {\tiny \bf \textsc{const\_dir}} at least} \colorbox{LIMIT}{\strut {\tiny \bf \textsc{limit}} 3000} strawberries and \colorbox{LIMIT}{\strut {\tiny \bf \textsc{limit}} 15000} raspberries.
        };

        \node [draw, align=left, text width=0.95\columnwidth, below=of INPUT] (OUTPUT) {
            A berry picker must pick <CONST\_DIR> at least </CONST\_DIR> <LIMIT> 3000 </LIMIT> strawberries and <LIMIT> 15000 </LIMIT> raspberries.
        };

        \draw [-Stealth] (INPUT.south) -- (OUTPUT.north);
    \end{tikzpicture}
    \caption{An example of named entity-based augmentation. In the input problem description, the named entities are encapsulated inside XML-like start and end tags of their respective types.}
    \label{fig:named_entity_tagging_example}
\end{figure}
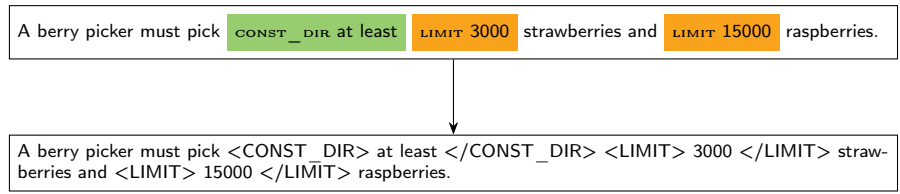

We use XML-like start and end tags to highlight the named entities in the input. We create XML-like tags for all named entity tags and encapsulate each entity in the input within these tags. Fig.~\ref{fig:named_entity_tagging_example} shows an example of this augmentation.

\subsubsection{Output.}
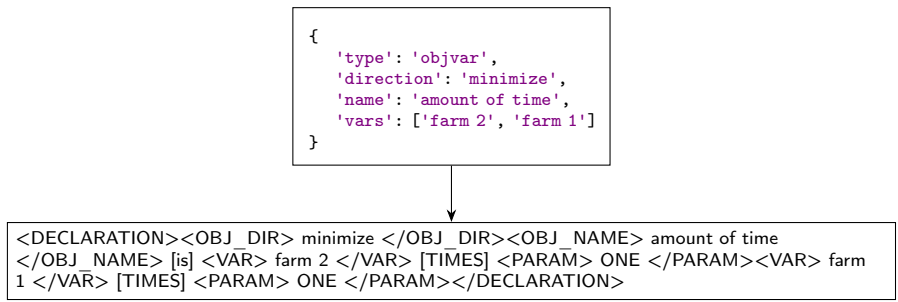
\begin{figure}[t]
    \centering
    \begin{tikzpicture}[font=\sffamily\scriptsize]
        \node [draw, text width=4cm, align=center] (JSON) {
        \begin{minted}[tabsize=2, autogobble, bgcolor=white]{bash}
{
    'type': 'objvar',
    'direction': 'minimize',
    'name': 'amount of time',
    'vars': ['farm 2', 'farm 1']
}
            \end{minted}
        };

        \node [draw, text width=0.95\columnwidth, align=left, below=of JSON, below=0.75cm] (TEXT) {
        <DECLARATION><OBJ\_DIR> minimize </OBJ\_DIR><OBJ\_NAME> amount of time </OBJ\_NAME> [is] <VAR> farm 2 </VAR> [TIMES] <PARAM> ONE </PARAM><VAR> farm 1 </VAR> [TIMES] <PARAM> ONE </PARAM></DECLARATION>
        };

        \draw [-Stealth] (JSON.south) -- (TEXT.north);
    \end{tikzpicture}
    \caption{The objective and constraints are converted to the XML format to form the intermediate representations.}
    \label{fig:output_json_to_xml}
\end{figure}

We follow a two-stage approach for generating the canonical representation of a linear programming problem. We use BART to generate an intermediate representation, which is then parsed into a canonical representation for evaluation. The intermediate representation consists of a set of declarations in an XML format, where each declaration corresponds to either an objective or a constraint. Fig.~\ref{fig:output_json_to_xml} shows an example of converting an objective to the intermediate representation.

The canonical form always minimizes the objective. In the case of a maximization objective, the sign of each objective parameter is inverted. Similarly, the inequality constraints are always assumed to have a ``$\leq$'' operator.  In the case of a ``$\geq$'' operator, the sign of each constraint parameter is inverted.

\subsubsection{Model.}
We use BART with the copy mechanism for our experiments \citep{lewis-etal-2020-bart, see-etal-2017-get}. We found that the copy mechanism had a small impact on the performance when the input problem description was augmented to highlight the named entities. Furthermore, we add new tokens, corresponding to the XML tags for the named entity tags, to the tokenizer and initialize their weights randomly at the time of training. At inference, we use greedy decoding to generate the output.

\section{Experiments}
\subsection{Dataset}
We use the NL4Opt Generation Dataset for our experiments \citep{ramamonjison2022augmenting}.\footnote{Available at \href{https://github.com/nl4opt/nl4opt-competition/}{github.com/nl4opt/nl4opt-competition}} This dataset consists of 1101 examples, divided into the train, dev, and test splits composed of 713, 99, and 289 examples, respectively. Each example consists of a linear programming word problem, labeled semantic entities, and the order mapping of named variables. These problems are from advertising, investment, sales, production, science, and transportation domains. The training split consists of problems from the first three domains. The dev and test splits contain problems from all six domains to evaluate the model's ability to generalize for domains it has not seen during training.

\subsection{Training Details}
The PyTorch transformers library implementation of base and large versions of BART was used \citep{wolf-etal-2020-transformers}. We use the AdamW optimizer with a learning rate of $5 \times 10^{-5}$ and a weight decay of $10^{-5}$ and a batch size of 16 with gradient accumulation for two steps for BART-large and a batch size of 32 with no gradient accumulation for BART-base \citep{loshchilov2017decoupled}. The models are fine-tuned for 400 epochs and are evaluated at the end of every epoch. We use a learning rate schedule with a linear warm-up for the first five epochs and a linear decay after that. The experiments are run on one A100 40 GB GPU.

\subsection{Evaluation Metrics}
The model is evaluated on declaration-level accuracy, where we refer to the objective or a constraint as a declaration \citep{ramamonjison2022augmenting}. The accuracy is defined as
\begin{equation}
    \textrm{Accuracy} = 1 - \frac{\sum_{i = 1}^{N} FP_i + FN_i}{\sum_{i = 1}^{N} D_i}
\end{equation}
where $N$ is the number of test examples. For $i^{th}$ example, $D_i$ is the number of ground truth declarations, $FP_i$ is the number of non-matched predicted declarations and $FN_i$ is the number of excess ground truth declarations.

\subsection{Results}
Table \ref{tab:results} shows the accuracy achieved by our approach. We initialize the training using five different seeds and report the best and mean accuracy achieved by our models on the test set. Our approach achieves the best accuracy among approaches that use \stos{} models, like BART, and fine-tune them on the generation dataset.

\begin{table}[t]
    \centering
    \caption{Accuracy achieved by our approach on the test set and a comparison with the existing approaches. For our approach, we report the best and mean (shown in brackets) accuracy values. The values for our approach are with greedy decoding.}
    \label{tab:results}
    \begin{small}
    \begin{sc}
    \setlength{\tabcolsep}{5pt}
    \begin{tabular}{ll}
    \toprule
    Approach & Accuracy \\
    \midrule
    \citet{Jang2022TagEA} & 0.878 \\
    \citet{ning2023novel} & 0.867 \\
    \citet{He2022LinearPW} & 0.780 \\
    \midrule
    \textbf{Our Approach} & \\
    w/ BART-base & 0.843 (0.817 $\pm$ 0.018) \\
    w/ BART-large & 0.920 (0.904 $\pm$ 0.011) \\
    \bottomrule
    \end{tabular}
    \end{sc}
    \end{small}
\end{table}




\subsection{Ablation Study}
\label{sec:ablation_study}
In this section, we discuss the impact of named entity augmentation, copy mechanism, and model size on the performance of the model on the generation task. We experiment with BART-base and BART-large as the pre-trained models and fine-tune the model by removing named entity augmentation and copy mechanism one at a time. We fix the other hyperparameters and initialize the training with five different seed values and report the max and mean accuracy. The results of the ablation study are shown in Table \ref{tab:ablation_study}.

\begin{table}[t]
    \centering
    \caption{Results of ablation study. We report the best and mean accuracy values achieved by the models with greedy decoding on the test set. Here, NEA and CM stand for named entity augmentation and copy mechanism, respectively.}
    \label{tab:ablation_study}
    \begin{small}
    \begin{sc}
    \setlength{\tabcolsep}{5pt}
    \begin{tabular}{lll}
    \toprule
    {} & BART-base & BART-large \\
    \midrule
    w/o CM & 0.827 (0.813 $\pm$ 0.010)  & 0.909 (0.899 $\pm$ 0.011) \\
    w/o NEA & 0.681 (0.659 $\pm$ 0.017) & 0.874 (0.838 $\pm$ 0.021) \\
    w/o NEA and CA & 0.666 (0.654 $\pm$ 0.008) & 0.867 (0.846 $\pm$ 0.016) \\
    \bottomrule
    \end{tabular}
    \end{sc}
    \end{small}
\end{table}

It can be observed from these results that the named entity augmentation results in an improvement with both the base and large versions of BART. This shows that highlighting named entities in the input text helps the model generate output formulations correctly. The copy mechanism also improves the performance by a couple of points. Furthermore, it should be noted that the performance improves significantly by adding named entity augmentation and copy mechanism on top of BART-base, making its performance comparable to the vanilla version BART-large. These results indicate that the need for a larger architecture may be offset by using data pre-processing and specialized architectures.

\section{Datasets without Labeled Named Entities}
In most applications, labeled named entity information is unavailable with the datasets. For these datasets, existing named entity recognition systems may be used to identify them. The state-of-the-art systems achieve >90\% accuracy on the CoNLL 2003 and NL4Opt NER tasks \citep{tjong-kim-sang-de-meulder-2003-introduction, ramamonjison2022augmenting, wang-etal-2021-automated, He2022LinearPW}. In this section, we investigate whether using these systems to label named entities can be useful for formulation generation.

\subsection{Noisy Named Entities}
Existing named entity recognition systems can label named entities with a certain accuracy. Errors from these systems fall into one of the following buckets:
\begin{enumerate}
    \itemsep0em
    \item A named entity span is missed by the system.
    \item A named entity span is identified correctly but labeled incorrectly.
    \item An excess named entity span is identified by the system.
\end{enumerate}

To simulate this behavior, for a fraction of labeled spans, $p$, we either drop the span, mislabel it or change its start and end positions, with an equal mix. We ensure that no two spans overlap. For our experiments, two datasets with noisy named entities are generated for $p = 0.2$ and $p = 0.5$. To quantify the extent of the noise in the named entity tags, we use the micro-averaged F1 score to compare the named entities in the generated and original datasets. Table \ref{tab:noise_f1} shows these scores, and Fig.~\ref{fig:noisy_ne_example} shows an example from the noisy datasets.

\begin{table}[t]
    \centering
    \caption{Comparison of noisy named entities with the original ground-truth labels (micro-averaged F1).}
    \label{tab:noise_f1}
    \begin{small}
    \begin{sc}
    \setlength{\tabcolsep}{5pt}
    \begin{tabular}{llll}
    \toprule
    $p$ & Train & Dev & Test \\
    \midrule
    0.2 & 0.8333 & 0.8397 & 0.8353 \\
    0.5 & 0.5783 & 0.5742 & 0.5822 \\
    \bottomrule
    \end{tabular}
    \end{sc}
    \end{small}
\end{table}

\begin{figure}[t]
    \centering
    \begin{tikzpicture}[font=\scriptsize\sffamily]
    \node [text width=0.98\textwidth] {A berry picker must pick \colorbox{CONSTDIR}{\strut {\tiny \textsc{const\_dir}} at least} \colorbox{LIMIT}{\strut {\tiny \textsc{limit}} 3000} strawberries and \colorbox{LIMIT}{\strut {\tiny \textsc{limit}} 15000} raspberries. He visits two farms. For each hour at \colorbox{VAR}{\strut {\tiny \textsc{var}} farm 1} he spends, \colorbox{PARAM}{\strut {\tiny \textsc{param}} he can} pick \colorbox{PARAM}{\strut {\tiny \textsc{param}} 50} strawberries and \colorbox{PARAM}{\strut {\tiny \textsc{param}} 300} raspberries. For each hour at farm 2 he spends, he can catch 70 strawberries and \colorbox{PARAM}{\strut {\tiny \textsc{param}} 200} raspberries. How many \colorbox{OBJDIR}{\strut {\tiny \textsc{obj\_dir}} hours} should he spend at each farm to \colorbox{OBJDIR}{\strut {\tiny \textsc{obj\_dir}} minimize} the \colorbox{OBJNAME}{\strut {\tiny \textsc{obj\_name}} amount of time} he spends at both farms?};
    \end{tikzpicture}

    \caption{An example of the input from the noisy dataset generated using $p = 0.5$. See Fig.~\ref{fig:overview} for the ground truth named entity spans and their labels.}
    \label{fig:noisy_ne_example}
\end{figure}

\subsection{Results}
We train and evaluate our approach on the noisy datasets generated through the process mentioned in the previous section. Table \ref{tab:noisy_dataset_results} shows the results of this experiment for $p = 0.2$ and $p = 0.5$. It can be observed that the named entity augmentation improves the performance even with noisy named entity tagging compared to vanilla BART models (See Section \ref{sec:ablation_study} for vanilla BART results). BART-base is affected more by the noise compared to BART-large. This is expected as BART-base also benefits more from named entity-based augmentation.

\begin{table}[t]
    \centering
    \caption{Accuracy achieved by our approach on the datasets with noisy named entities. A lower value of $p$ implies more accurate named entity labeling.}
    \label{tab:noisy_dataset_results}
    \begin{small}
    \begin{sc}
    \setlength{\tabcolsep}{5pt}
    \begin{tabular}{lll}
    \toprule
    $p$ & BART-base & BART-large \\
    \midrule
    0.2 & 0.744 (0.734 $\pm$ 0.012) & 0.896 (0.885 $\pm$ 0.011) \\
    0.5 & 0.722 (0.704 $\pm$ 0.011) & 0.890 (0.870 $\pm$ 0.016) \\
    \bottomrule
    \end{tabular}
    \end{sc}
    \end{small}
\end{table}

\section{Conclusion}
In this paper, we proposed a novel approach based on highlighting named entities in the input text to generate the mathematical formulation of linear programming word problems. Our approach produced the highest accuracy among all submissions to the generation track of the NL4Opt Competition. We also found that the ``all-at-once'' strategy for generation performs better than generating the objective or a constraint at a time with the named entity-based augmentation. Lastly, we also showed that applications without labeled named entities might use named entity-based augmentation by first labeling the named entities using an existing system. We believe that this augmentation may prove useful in other natural language processing tasks.

\subsubsection{Acknowledgements.}
We thank Prof. Shaloo Rakheja (University of Illinois Urbana-Champaign) for providing computing resources for this work. This work also utilizes resources supported by the National Science Foundation's Major Research Instrumentation program, grant \#1725729, as well as the University of Illinois at Urbana-Champaign.

\bibliographystyle{splncs04nat}
\renewcommand{\bibsection}{\section*{References}}
\bibliography{anthology, custom}

\begin{thebibliography}{16}
\providecommand{\natexlab}[1]{#1}
\providecommand{\url}[1]{\texttt{#1}}
\providecommand{\urlprefix}{URL }
\expandafter\ifx\csname urlstyle\endcsname\relax
  \providecommand{\doi}[1]{doi:\discretionary{}{}{}#1}\else
  \providecommand{\doi}{doi:\discretionary{}{}{}\begingroup
  \urlstyle{rm}\Url}\fi

\bibitem[{Beairsto et~al.(2021)Beairsto, Tian, Zheng, Zhao, and
  Hong}]{beairsto2021identifying}
Beairsto, J., Tian, Y., Zheng, L., Zhao, Q., Hong, J.: Identifying locations
  for new bike-sharing stations in glasgow: an analysis of spatial equity and
  demand factors. Annals of GIS pp. 1--16 (2021)

\bibitem[{Bitran and Caldentey(2016)}]{bitran2016overview}
Bitran, G., Caldentey, R.: An overview of pricing models for revenue
  management. IEEE Engineering Management Review \textbf{44}(4), 134--134
  (2016)

\bibitem[{He et~al.(2022)He, Mamatha, Vignesh, Kumar, and
  Uppal}]{He2022LinearPW}
He, J., Mamatha, N., Vignesh, S., Kumar, D., Uppal, A.: Linear programming word
  problems formulation using ensemblecrf ner labeler and t5 text generator with
  data augmentations. ArXiv \textbf{abs/2212.14657} (2022)

\bibitem[{Jang(2022)}]{Jang2022TagEA}
Jang, S.: Tag embedding and well-defined intermediate representation improve
  auto-formulation of problem description. ArXiv \textbf{abs/2212.03575} (2022)

\bibitem[{Karmarkar(1984)}]{karmarkar1984new}
Karmarkar, N.: A new polynomial-time algorithm for linear programming. In:
  Proceedings of the sixteenth annual ACM symposium on Theory of computing, pp.
  302--311 (1984)

\bibitem[{Lewis et~al.(2020)Lewis, Liu, Goyal, Ghazvininejad, Mohamed, Levy,
  Stoyanov, and Zettlemoyer}]{lewis-etal-2020-bart}
Lewis, M., Liu, Y., Goyal, N., Ghazvininejad, M., Mohamed, A., Levy, O.,
  Stoyanov, V., Zettlemoyer, L.: {BART}: Denoising sequence-to-sequence
  pre-training for natural language generation, translation, and comprehension.
  In: Proceedings of the 58th Annual Meeting of the Association for
  Computational Linguistics, pp. 7871--7880, Association for Computational
  Linguistics, Online (Jul 2020), \doi{10.18653/v1/2020.acl-main.703},
  \urlprefix\url{https://aclanthology.org/2020.acl-main.703}

\bibitem[{Loshchilov and Hutter(2017)}]{loshchilov2017decoupled}
Loshchilov, I., Hutter, F.: Decoupled weight decay regularization. arXiv
  preprint arXiv:1711.05101  (2017)

\bibitem[{Ma et~al.(2016)Ma, Qin, Xu, and Zou}]{ma2016research}
Ma, Y., Qin, X., Xu, J., Zou, X.: Research on pricing method of public bicycle
  service: A case study in guangzhou. In: 2016 IEEE International Conference on
  Service Operations and Logistics, and Informatics (SOLI), pp. 156--161, IEEE
  (2016)

\bibitem[{Nash(2000)}]{nash2000dantzig}
Nash, J.C.: The (dantzig) simplex method for linear programming. Computing in
  Science \& Engineering \textbf{2}(1), 29--31 (2000)

\bibitem[{Ning et~al.(2023)Ning, Liu, Qin, Xiao, Xue, Huang, Liu, Chen, and
  Wu}]{ning2023novel}
Ning, Y., Liu, J., Qin, L., Xiao, T., Xue, S., Huang, Z., Liu, Q., Chen, E.,
  Wu, J.: A novel approach for auto-formulation of optimization problems. arXiv
  preprint arXiv:2302.04643  (2023)

\bibitem[{Ramamonjison et~al.(2022)Ramamonjison, Li, Yu, He, Rengan,
  Banitalebi-Dehkordi, Zhou, and Zhang}]{ramamonjison2022augmenting}
Ramamonjison, R., Li, H., Yu, T.T., He, S., Rengan, V., Banitalebi-Dehkordi,
  A., Zhou, Z., Zhang, Y.: Augmenting operations research with auto-formulation
  of optimization models from problem descriptions. arXiv preprint
  arXiv:2209.15565  (2022)

\bibitem[{See et~al.(2017)See, Liu, and Manning}]{see-etal-2017-get}
See, A., Liu, P.J., Manning, C.D.: Get to the point: Summarization with
  pointer-generator networks. In: Proceedings of the 55th Annual Meeting of the
  Association for Computational Linguistics (Volume 1: Long Papers), pp.
  1073--1083, Association for Computational Linguistics, Vancouver, Canada (Jul
  2017), \doi{10.18653/v1/P17-1099},
  \urlprefix\url{https://aclanthology.org/P17-1099}

\bibitem[{Tao et~al.(2020)Tao, Pleau, Akridge, Fradet, Grondin, Laughlin,
  Miller, and Shoemaker}]{tao2020analytics}
Tao, D.Q., Pleau, M., Akridge, A., Fradet, O., Grondin, F., Laughlin, S.,
  Miller, W., Shoemaker, L.: Analytics and optimization reduce sewage overflows
  to protect community waterways in kentucky. INFORMS Journal on Applied
  Analytics \textbf{50}(1), 7--20 (2020)

\bibitem[{Tjong Kim~Sang and
  De~Meulder(2003)}]{tjong-kim-sang-de-meulder-2003-introduction}
Tjong Kim~Sang, E.F., De~Meulder, F.: Introduction to the {C}o{NLL}-2003 shared
  task: Language-independent named entity recognition. In: Proceedings of the
  Seventh Conference on Natural Language Learning at {HLT}-{NAACL} 2003, pp.
  142--147 (2003), \urlprefix\url{https://aclanthology.org/W03-0419}

\bibitem[{Wang et~al.(2021)Wang, Jiang, Bach, Wang, Huang, Huang, and
  Tu}]{wang-etal-2021-automated}
Wang, X., Jiang, Y., Bach, N., Wang, T., Huang, Z., Huang, F., Tu, K.:
  Automated concatenation of embeddings for structured prediction. In:
  Proceedings of the 59th Annual Meeting of the Association for Computational
  Linguistics and the 11th International Joint Conference on Natural Language
  Processing (Volume 1: Long Papers), pp. 2643--2660, Association for
  Computational Linguistics, Online (Aug 2021),
  \doi{10.18653/v1/2021.acl-long.206},
  \urlprefix\url{https://aclanthology.org/2021.acl-long.206}

\bibitem[{Wolf et~al.(2020)Wolf, Debut, Sanh, Chaumond, Delangue, Moi, Cistac,
  Rault, Louf, Funtowicz, Davison, Shleifer, von Platen, Ma, Jernite, Plu, Xu,
  Le~Scao, Gugger, Drame, Lhoest, and Rush}]{wolf-etal-2020-transformers}
Wolf, T., Debut, L., Sanh, V., Chaumond, J., Delangue, C., Moi, A., Cistac, P.,
  Rault, T., Louf, R., Funtowicz, M., Davison, J., Shleifer, S., von Platen,
  P., Ma, C., Jernite, Y., Plu, J., Xu, C., Le~Scao, T., Gugger, S., Drame, M.,
  Lhoest, Q., Rush, A.: Transformers: State-of-the-art natural language
  processing. In: Proceedings of the 2020 Conference on Empirical Methods in
  Natural Language Processing: System Demonstrations, pp. 38--45, Association
  for Computational Linguistics, Online (Oct 2020),
  \doi{10.18653/v1/2020.emnlp-demos.6},
  \urlprefix\url{https://aclanthology.org/2020.emnlp-demos.6}

\end{thebibliography}

\end{document}